\documentclass[conference]{IEEEtran}
\IEEEoverridecommandlockouts
\usepackage{cite}
\usepackage{amsmath,amssymb,amsfonts}
\usepackage{algorithmic}
\usepackage{algorithm}

\usepackage{graphicx}
\usepackage{textcomp}
\usepackage{xcolor}
\usepackage{tikz,bm,balance}
\usepackage{pgfplots}
\usepackage{hyperref}
\usetikzlibrary{patterns}
\usepackage{subcaption}
\def\BibTeX{{\rm B\kern-.05em{\sc i\kern-.025em b}\kern-.08em
    T\kern-.1667em\lower.7ex\hbox{E}\kern-.125emX}}

\definecolor{darkgreen}{rgb}{0., 0.6, 0.}

\begin{document}

\title{Jammer classification with Federated Learning\\
\thanks{This work has been partially supported by the National Science Foundation under Award ECCS-1845833.}
}


\author{\IEEEauthorblockN{Peng Wu, Helena Calatrava, Tales Imbiriba, Pau Closas}
\IEEEauthorblockA{\textit{Northeastern University} \\
\textit{Dept. of Electrical \& Computer Eng.}\\
 Boston, MA (USA) \\
\{wu.p, calatrava.h, talesim, closas\}@northeastern.edu}
}

\maketitle

\begin{abstract}
Jamming signals can jeopardize the operation of GNSS receivers until denying its operation. Given their ubiquity, jamming mitigation and localization techniques are of crucial importance, for which jammer classification is of help.
Data-driven models have been proven useful in detecting these threats, while their training using crowdsourced data still poses challenges when it comes to private data sharing.
This article investigates the use of federated learning to train jamming signal classifiers locally on each device, with model updates aggregated and averaged at the central server. This allows for privacy-preserving training procedures that do not require centralized data storage or access to client local data. The used framework FedAvg is assessed on a dataset consisting of spectrogram images of simulated interfered GNSS signal. Six different jammer types are effectively classified with comparable results to a fully centralized solution that requires vast amounts of data communication and involves privacy-preserving concerns.
\end{abstract}


\section{Introduction}\label{sec:introduction}
GNSS jamming signals are L-band spectrum interferences that can overpower a GNSS receiver until denying its operation \cite{amin2016vulnerabilities,morton2021position}. A wide variety of jammers can be found in the online market at very cheap prices, which makes human-made intentional jamming signals a threat \cite{borio2016impact,Ferre2020}. In addition, signals do not need to be malicious to have a jamming effect, where multiple examples exist of legitimate waveforms that can pose a threat to GNSS receivers, such as continuous wave (CW) interferences produced by damaged electronics and signals emitted by Distance Measurement Equipment (DME) technology conceived for aircraft navigation~\cite{li_dual-domain_2019}.
Jamming sources are placed on Earth or, in the case of drone jammers, in the proximity of the Earth's surface. As a consequence of the path-loss attenuation given by the large distance between Earth and GNSS satellites, jamming interferences are received with remarkably higher power than the useful GNSS signal, which can lead to performance disruption in areas with a radius of several kilometers~\cite{Mitch2011}. In the literature, it is suggested that jamming is the main cause of GNSS-based service outages~\cite{Ferre2019} and, consequently, we consider that protection against this kind of attack is a desirable feature in GNSS receivers \cite{dovis2015gnss,thombre2018gnss}.

Jammer classification can be of great help to classical Interference Classification (IC) techniques, which are formulated as an estimation problem where the jamming signal is detected and estimated, often with a parametric model~\cite{closas_afreshlook}. As the aim of these techniques is to first reconstruct the interference, the knowledge of its type or class is key to speed up the algorithm. For instance, if knowing that a CW interference is threatening a receiver, it would only be required to estimate the interference central frequency in order to reconstruct its waveform and implement an IC measure. It is also relevant to highlight that by performing jamming classification, the task of detection is explicitly taken care of. In the vast majority of previous GNSS studies regarding protection against jamming interferences, the focus is on its detection~\cite{youness2020}, mitigation~\cite{borio_hubers_2018}, and localization~\cite{Strii2017CrowdsourcingGJ}. Nevertheless, and according to~\cite{Ferre2019}, little effort had been dedicated to the classification of jamming signals until recent publications, besides some work in the context of radar systems such as the Machine Learning (ML) jamming prediction algorithm proposed in~\cite{lee_jamming_2020}. In~\cite{Ferre2019}, they propose a Support Vector Machine (SVM) and a Convolutional Neural Network (CNN) based classifiers for the purpose of jammer classification, which they treat as an image classification problem. They suggest that with a small amount of training data, it is possible to achieve classification accuracy above 90\%, being the accuracy of jamming detection close to 99\%. The use of multivariate time-series approaches can also lead to an increase in classification accuracy in jammer classification techniques, according to the work presented in~\cite{voigt_classification_2021}, which makes use of state-of-the-art ML techniques.

\begin{figure*}[ht]
\centerline{\includegraphics[width=0.95\textwidth]{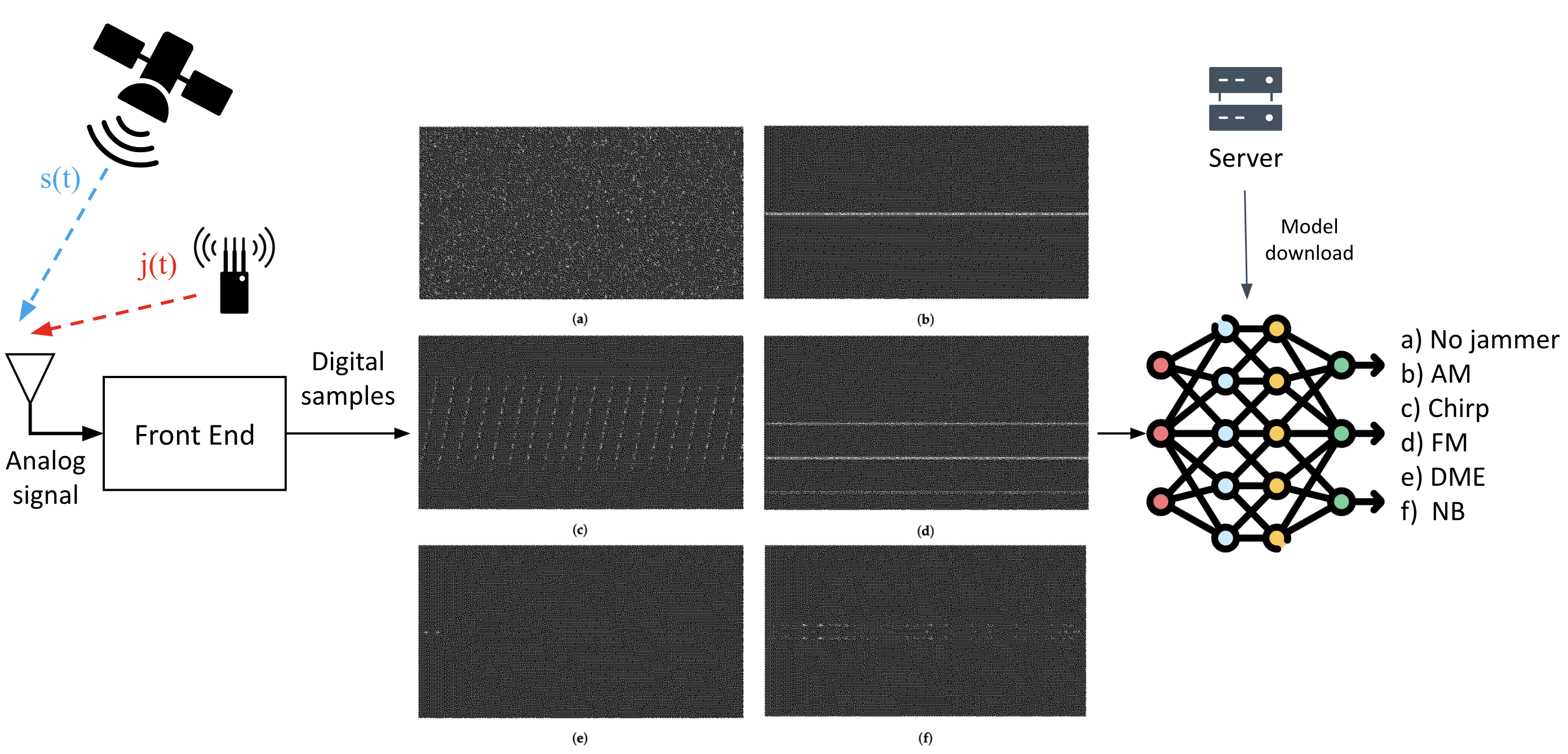}
}
\caption{System diagram of the considered jamming signal classification: a receiver downloads a pretrained model from the server, which can be either $i)$ trained on locally available data and sent back to the server for fusion with other models; or $ii)$ used to perform jamming classification results based on local data. Monochrome spectrogram images of the six jammer types available in the used dataset~\cite{Ferre2019} are shown, namely (b) Amplitude Modulated (AM), (c) chirp, (d) Frequency Modulated (FM), (e) Pulsed or Distance Measurement Equipment (DME) and (f) Narrow band (NB) jammers. Class (a) corresponds to clean signal (no interference).}
\label{fig:jammer_detect}
\end{figure*}

%
Most studies on GNSS integrity rely on synthetic data since data collection in the presence of jamming signals can represent a tremendous effort. This is especially so if different interference types and received power values are desired. Besides recreating effects such as the ones introduced by multipath reflections can be difficult.
The use of real GNSS interfered data is, however, of great interest when it comes to training data-driven classifiers, as well as  effectively assessing their performance.
An option to collect real GNSS data is to resort to traditional crowdsourcing approaches, where clients record data and share it with a central unit that is in charge of training the classifier. Nevertheless, crowdsourcing has some concerns about user privacy, as it requires that the users involved in the scenario send their data directly to a centralized server. Aimed at solving this limitation, Federated Learning (FL) has recently attracted great interest due to its privacy-protecting nature and the efficient use of resources by harnessing the processing power of edge devices~\cite{niknam2020federated}.
FL is a promising solution that enables many clients to jointly train machine learning models while maintaining local data decentralization. Collaboration between users in distributed scenarios has been proven useful in GNSS interference management tasks~\cite{nicola2020collaborative}. With FL, instead of exchanging data and conducting centralized training, each party sends its model to the server, which updates a joint model and sends the global model back to the parties.
Since their original data is not exposed, FL is an effective way to address privacy issues \cite{mcmahan2017communication}.
%

\begin{figure}
\centerline{\includegraphics[width=0.48\textwidth]{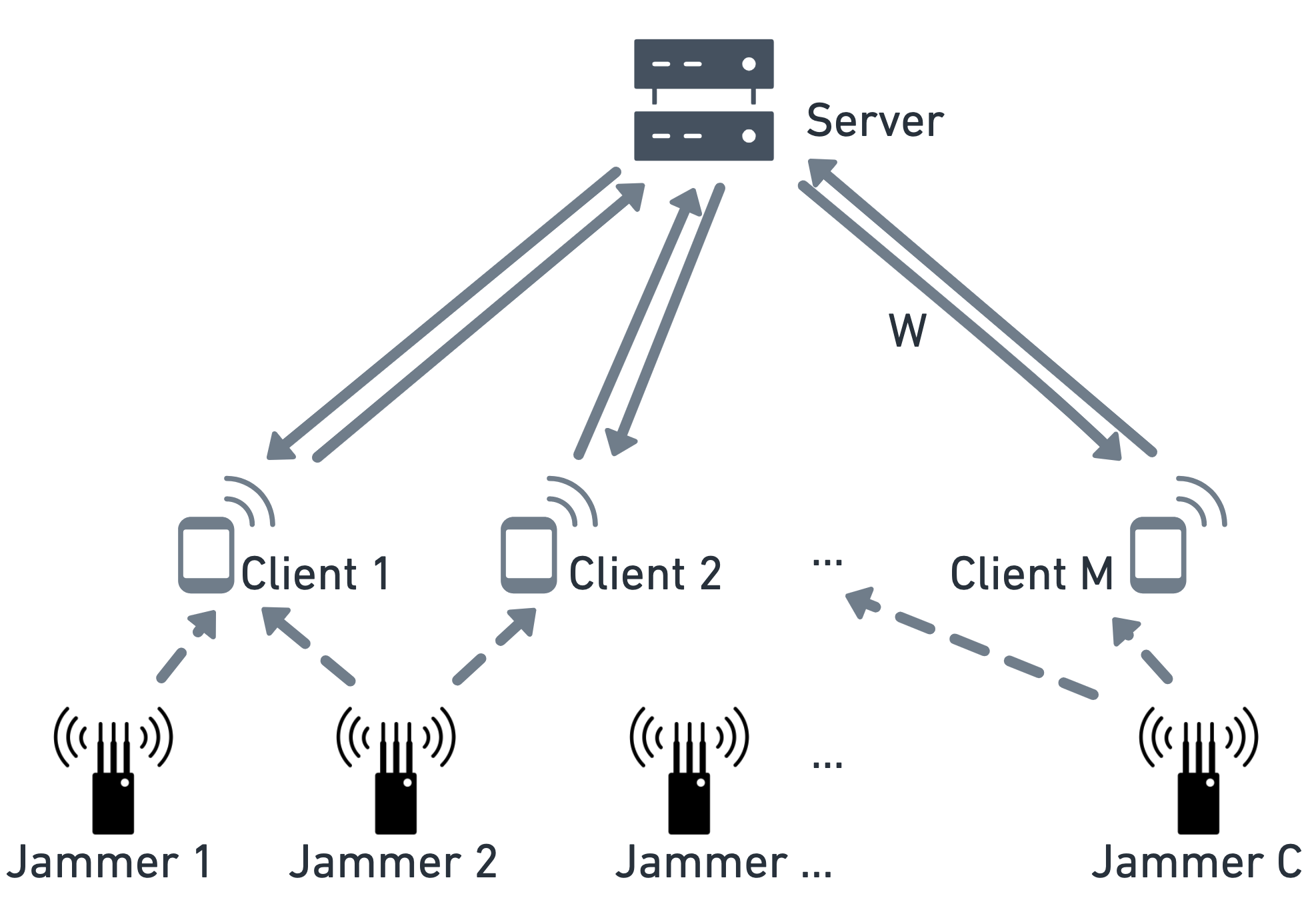}
}
\caption{Federated learning framework to train jamming signal classifiers: $M$ collaborative clients receive the parameters of the classifier from a server; these clients retrain the model based on local data; and upload their updated classifier to the server in charge of fusing the results. This process does not require exchange of actual data or positions from the clients, thus preserving their privacy.}
\label{fig:fedjammer}
\end{figure}

In this paper, we aim at training jamming signal classifiers using privacy-preserving strategies that can cope with crowdsourcing data collection strategies. Our overall goal is to obtain a Neural Network (NN) based global model capable of classifying different jamming signals as depicted in Figure~\ref{fig:jammer_detect}. To preserve client privacy while leveraging crowdsourcing data collection strategies we exploit FL approaches as shown in Figure~\ref{fig:fedjammer} where model parameters are shared with clients allowing for local classification of jamming signals and avoiding data sharing. In the proposed framework we assume the possible existence of $C$ different jamming types while the FL approach is performed over a network with $M$ collaborative users. 
We study the FL-based jamming classifier under different data distribution scenarios. In the first scenario clients' data is independent and identically distributed (IID), that is, all clients observe a similar amount of instances from all $C$ classes. In the second, and more challenging, scenario clients observe data that is unbalanced towards different classes. 
Working with non-IID data poses several challenges that are common in realistic scenarios, given that not all clients have access to all available types of data. In the context of this work, this is the case when not all participating users observe the same classes of jammers. 

The remainder of this paper is organized as follows. Section~\ref{sec:system_model} provides a description of the satellite signal model and targeted jammer types. The used FL technique is derived in Section~\ref{sec:methodology}, while the experimental setup and results can be found in Section~\ref{sec:experiments}. Finally, Section~\ref{sec:conclusion} concludes the paper.  

\section{System Model}\label{sec:system_model}
For the purpose of this article, the analog baseband equivalent of the received GNSS signal can be modeled as
\begin{equation}\label{eq:signalmodel}
    r\left(t\right) = s\left(t\right) + j\left(t\right) + w\left(t\right),
\end{equation}
where $s\left(t\right)$ contains the useful GNSS satellite signals and $w\left(t\right)$ represents sources of randomness such as thermal noise, typically modeled as an additive white Gaussian noise (AWGN) process. The term $j\left(t\right)$ represents the signal waveform generated by a jamming source, as measured at the receiver. Several waveforms are possible for $j(t)$ depending on the type of jammer \cite{Ferre2020}. Accurate knowledge of $j(t)$ allows for prompt reaction to a jamming threat, either for its localization \cite{narding2023jammerlocaliz} or mitigation. Related to the latter, IC techniques aim to estimate the 
waveform of $j\left(t\right)$ so that it can be reconstructed and directly subtracted from $r\left(t\right)$. As it has been previously mentioned in Section~\ref{sec:introduction}, identifying the type of waveform of $j\left(t\right)$ can be useful for several purposes, including its reconstruction and mitigation through IC techniques.

Jammers can be classified according to several features such as the type of device by which they are broadcasted, their frequency spectrum, and their number of antennae \cite{borio2016impact}. In this paper, we are targeting the same jammer types as in~\cite{Ferre2019}, given that we are using the dataset they provide and using their results as a benchmark. The aim of our research is to use the FL technique explained in Section~\ref{sec:methodology} for the classification of the following jammer types. This classification is mostly performed according to their behavior in the frequency domain. 

\begin{enumerate}
  \item Amplitude Modulated (AM);
  \item Chirp;
  \item Frequency Modulated (FM);
  \item Pulsed or Distance Measurement Equipment (DME);
  \item Narrow Band (NB) jammers; and
  \item No interference.
\end{enumerate}

As in~\cite{Ferre2019}, we are not considering Wide Band (WB) jammers given that it is very difficult to detect their presence when analyzing spectrogram images. Jammers of types 1) to 5) have narrow spectrums which overpower the signal of interest buried in noise. We would also like to point out that the classification strategy, proposed in Section~\ref{sec:methodology}, is able to perform the detection task, as the absence of interference can be properly identified. The five waveform expressions of $j\left(t\right)$ for each jammer type presented in the list above can be found in~\cite{Ferre2019}, and are not explicitly used in both training or testing of the proposed FL solution. 

While AM and FM jammers target pre-fixed frequencies, other jammers such as chirp jammers sweep over different frequency bands. Consequently, feature extraction approaches based on spectral analysis, such as spectrograms, of the signals are suitable for distinguishing different jammer types. This is so since the short-time Fourier transform allows the time-frequency localization of the interference signal. 
%
In~\cite{Ferre2019}, they successfully approached jammer classification as an image classification problem, where spectrograms of the received signal $r(t)$ were treated as images. More precisely, the spectrograms are computed on the discrete-time version of $r(t)$ in \ref{eq:signalmodel}, which at an appropriate sampling rate $f_s=1/T_s$ would be modeled as $r[n] = s[n] + j[n] + w[n]$ where $t=n T_s$ for $n\in \mathbb{Z}$.

\section{Federated Learning Methodology}\label{sec:methodology}

A significant number of FL algorithms have been discussed in different areas \cite{li2021federatedLO, 9593115}, especially in the field of image classification. One \textit{de facto} approach for FL is Federated Averaging (FedAvg) \cite{mcmahan2017communication}, which fuses the model parameters by a weighted sum. 
According to previous studies~\cite{hsu2019measuring, li2019convergence}, the learning effectiveness of standard FL methods is compromised under non-IID data settings. 

In this section, we aim on leveraging FL strategies to learn a unique global model capable of making accurate predictions on data available to different clients. 
More precisely, we consider the setup depicted in Fig. \ref{fig:fedjammer}, where $M$ collaborative clients aim at training a global classification model (e.g. a neural network) such that class posteriors
\begin{equation}
    \mathbf{y}= \mathbf{h}(\mathbf{X} ; \bm{\omega})
\end{equation}
\noindent where $\mathbf{y}\in\mathbb{R}^C$ is the vector of class posteriors with elements $p(y=\ell|\mathbf{X})$, with $\ell\in \{\mathrm{AM,\, Chirp, FM, DME, NB, NO}\}$, $\mathbf{h}:\mathbf{X}\mapsto \mathbf{h}(\mathbf{X})$ is the NN classifier parameterized by $\bm{\omega}\in\mathbb{R}^{N_\omega}$, and $\mathbf{X}\in\mathbb{R}^{T_w\times N}$ is the spectrogram of the received GNSS signal $\mathbf{r}[n]$, see \cite{Ferre2019} for more details regarding the construction of the spectrogram data. 
In this contribution, we assume that the data $\mathcal D $ is composed of $M$ disjoint datasets $\mathcal D_i = \{\mathbf{y}^{(i)}_n , \mathbf{X}_n^{(i)}\}_{n=1}^{L_i}$, $i\in\{1,\ldots, M\}$.

Mathematically, the training process can be formulated as the minimization of a loss function:
\begin{equation}\label{eq:FL}
\min_{\bm{\omega}} \mathcal{L}(\bm{\omega}) \text{     where    } \mathcal{L}(\bm{\omega}) = \sum_{i=1}^{M} \mathcal{F}_i(\bm{\omega})
\end{equation}
where $\mathcal{L}(\bm{\omega})$ is the global loss functional, while $\mathcal{F}_i: \mathbb{R}^d\rightarrow \mathbb{R}, \bm{\omega} \mapsto \mathcal{F}_i(\bm{\omega})$ 
are local loss functions.

Among the different strategies to solve \eqref{eq:FL} we highlight the
conventional FL approach: FedAvg, which considers a single
global objective, along with other variants such as FedProx, \cite{li2020federated}, which consider adding a regularization term to the objective function, to prevent overfitting, and MOON \cite{li2021model}, which use a contrastive loss term to control the local model drifts away from the global model. However, our practical experience with these three methods, when applied to the jammer classification problem at hand, is that they perform very similarly with FedAvg presenting a slight advantage. Thus, we report results only for the FedAvg in our experiments in Section~\ref{sec:experiments}.

FedAvg is a \textit{de facto} approach for FL in which local models are trained locally. 
The clients then upload local trained model parameters to a cloud server, which is in charge of fusing it (e.g. weights in a neural network) to compute a unique global model. 
Using the cardinality of the local data $|\mathcal D_i|$ as a metric of model reliability, 
\eqref{eq:FL} is modified as:
\begin{equation}\label{eq:Fedavg}
\mathcal{L}(\bm{\omega}) = \sum_{i=1}^{M} \frac{N_i}{N} \mathcal{L}_i(\bm{\omega})
\end{equation}
where $\mathcal{L}_{i}(\bm{\omega})=\frac{1}{N_{i}} \sum_{n \in \mathcal{D}_{i}} f_{n}(\bm{\omega})$, $f_n(\bm{\omega})$ is the loss of the prediction using sample $n$ from the dataset $\mathcal{D}_{i}$.
$\mathcal{D}_{i}$ is the data partition for client $i$, $N_i$ is the number samples available to the $i$-th client in $\mathcal{D}_i$, and $N=N_1+\dots+N_M$ is the total number of data points. 

The optimization in~\eqref{eq:Fedavg} is solved iteratively through multiple rounds of local optimization in the clients and fusion in the server. First, at iteration $t$ each client updates its model parameters solving:
\begin{equation}\label{eq:local_gd}
    \bm{\omega}_{t+1}^i = \mathop{\arg\min}_{\bm{\omega}} \mathcal{L}_{i,t}(\bm{\omega})
\end{equation}
where the index $t$ in $\mathcal{L}_{i,t}(\bm{\omega})$ indicates that the local parameters were initialized using the fused global parameters, $\bm \omega_t$, from the previous iteration.
Secondly, a \textit{global update equation} is used, where the global model parameter is updated by averaging the locally updated models from each device as:
\begin{equation}\label{eq:aggregation}
    \bm{\omega}_{t+1} = \sum_{i=1}^{M} \frac{N_i}{N} \bm{\omega}_{t+1}^i
\end{equation}
where $\bm{\omega}_{t+1}$ is the updated global model, and the sum is over all locally updated models $\bm{\omega}_{t+1}^i$ from each device.
The implementation details of FedAvg are shown in Algorithm \ref{alg:FedAvg}.

\begin{algorithm}[tb]
  \caption{FedAvg Algorithm}
  \label{alg:FedAvg}
  \begin{algorithmic}
    \STATE \textbf{Input:} 
    number of clients $M$; the architecture of local models $\mathbf h$ with initial $\bm{\omega}_0$; local loss functions $\mathcal F_i$; data $\mathcal D = \{\mathcal{D}_1,\ldots,\mathcal{D}_M\}$; number of iterations $T$; number of local epochs $E$; learning rate $\eta$;
    \FOR{$t=1, \ldots, T$}
      \FOR{$i \in M$}
        \STATE $\bm{\omega}_{t+1}^i \gets$ solution of \eqref{eq:local_gd} using local data $\mathcal{D}_i$ for $E$ epochs with learning rate $\eta$ 
        \STATE Upload local model parameters $\bm{\omega}_{t+1}^i$ to server.
      \ENDFOR
      \STATE Update global model parameters $\bm{\omega}_{t+1}$ with equation \eqref{eq:aggregation} and send it to local clients.
    \ENDFOR
    \STATE \textbf{Output:} $\bm{\omega}_T$
  \end{algorithmic}
\end{algorithm}




\section{Experiments}\label{sec:experiments}

This section presents a set of experiments to show the applicability of federated learning to train, in a distributed manner, a jammer classifier able to achieve performances closer to those from a classifier trained on a centralized node with access to all local datasets. We first describe the dataset used, then how it is employed in a distributed learning scheme, how the model was configured, and finally the obtained results.

\subsection{Data preprocessing}
The dataset provided by the authors in~\cite{Ferre2019}, which is available in open access at \url{https://zenodo.org/record/3370934}, is used to conduct the following experiments. It contains $61800$ available \textit{.bmp} monochrome spectrogram images with $512\times 512$ pixel resolution, binary scale and $600$ DPI. To compute the spectrograms, simulated GNSS signal interfered by the aforementioned jammer types (see Section~\ref{sec:system_model}) is processed. In \cite{Ferre2019}, the authors used $6000$ images for training ($1000$ for each jammer class), $1800$ images for validation and $54000$ for testing.  

 In order to optimize computational resources and expedite the training process we performed some data preprocessing, approach that is usually employed in machine learning context. Particularly, in this paper, we utilized both the training and validation datasets (validation step is often omitted from the experimentation process, unless performing hyperparameter tuning), the combined dataset was then split into a $75\%$ train and $25\%$ test division. Additionally, to further enhance the process, image resolution was reduced from $512 \times 512$ to $256 \times 256$ pixels through the use of bilinear interpolation techniques. Additionally, once all the data was preprocessed, it was normalized in order to facilitate the training phase.



\subsection{Federated data setting}
Two different data settings were investigated. First, the case of an IID setting, wherein all clients received similar data distributions, that is, a similar amount of samples from each class. For the experiments, we uniformly split the data into $20$, $30$, and $40$ clients, in order to examine how client numbers may influence the results. 
This split resulted in approximately $65$, $43$, and $32$ samples per client for $20$, $30$, and $40$ clients, respectively.  

\begin{figure}
\centerline{\includegraphics[width=0.48\textwidth]{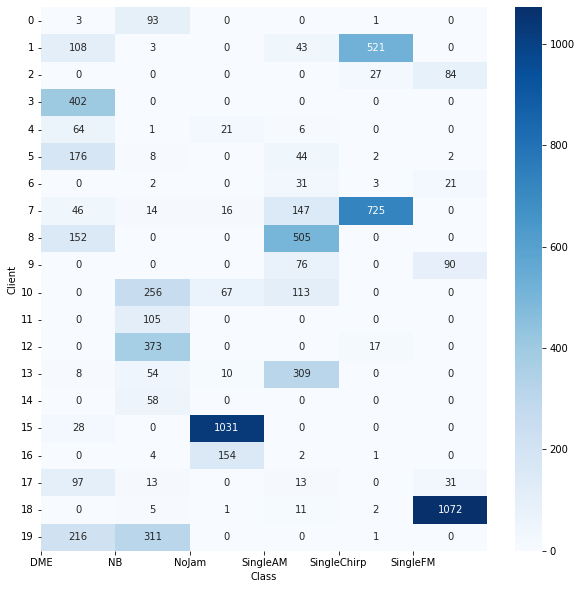}
}
\caption{Number of data points per classes for each of the $M=20$ clients.}
\label{fig:data_20_dir}
\end{figure}

The second set of experiments was for a non-IID setting, where the focus is on having an unbalanced distribution of the class labels for training.  
To generate non-IID splits of the dataset, we followed the approach in~\cite{li2021federatedLO}, where client data is sampled using a Dirichlet distribution. 
%
Specifically, for a given client $i$, we defined the probability of sampling data from a label $j \in \{1,\dots,C\}$ as the vector $(p_{i,1},\dots,p_{i,C}) \sim 
\operatorname{Dir}(\bm{\beta})$, where $\operatorname{Dir}(\cdot)$ denotes the Dirichlet distribution and $\bm{\beta} = (\beta_{i,1}, \dots, \beta_{i,C})^\top$ is the concentration vector parameter. 
%
The advantage of this approach is that the imbalance level can be flexibly changed by adjusting the concentration parameter $\beta_{i,j}$. In this paper, the concentration parameter $\beta_{i,j}$ is set to a relatively small value of $0.1$, allowing for a more unbalanced partitioning. This is evident when inspecting the distribution of data points among clients, where many clients only contain a few labels. This can be observed in Figure \ref{fig:data_20_dir}, providing a snapshot of the number of samples per class for each client when $M=20$ clients. This leads to an unequal partition of the data, with some clients containing a disproportionately large or small percentage of certain class labels.

\subsection{Model setting}
The authors in \cite{morales2019jammer} employed a convolutional neural network (CNN) for the task of training a classifier based on the full dataset $\mathbf{D}$. This solution becomes the baseline in our results, where the same CNN architecture is considered, while it is trained using the FL framework described earlier. In particular, the architecture of the CNN consisted of one convolutional layer, one pooling layer, and one fully connected layer with a ReLU activation function. The convolution layer utilized $16$ filters of size $12 \times 12 \times 1$, with a learning rate of $0.01$, and an SGD optimizer \cite{ruder2016overview} was used. The last layer is softmax layer to produce classification results. Plus, Cross-entropy is used for cost function.


\subsection{Results}\label{sec:results}


\begin{figure}
\centerline{\includegraphics[width=0.50\textwidth]{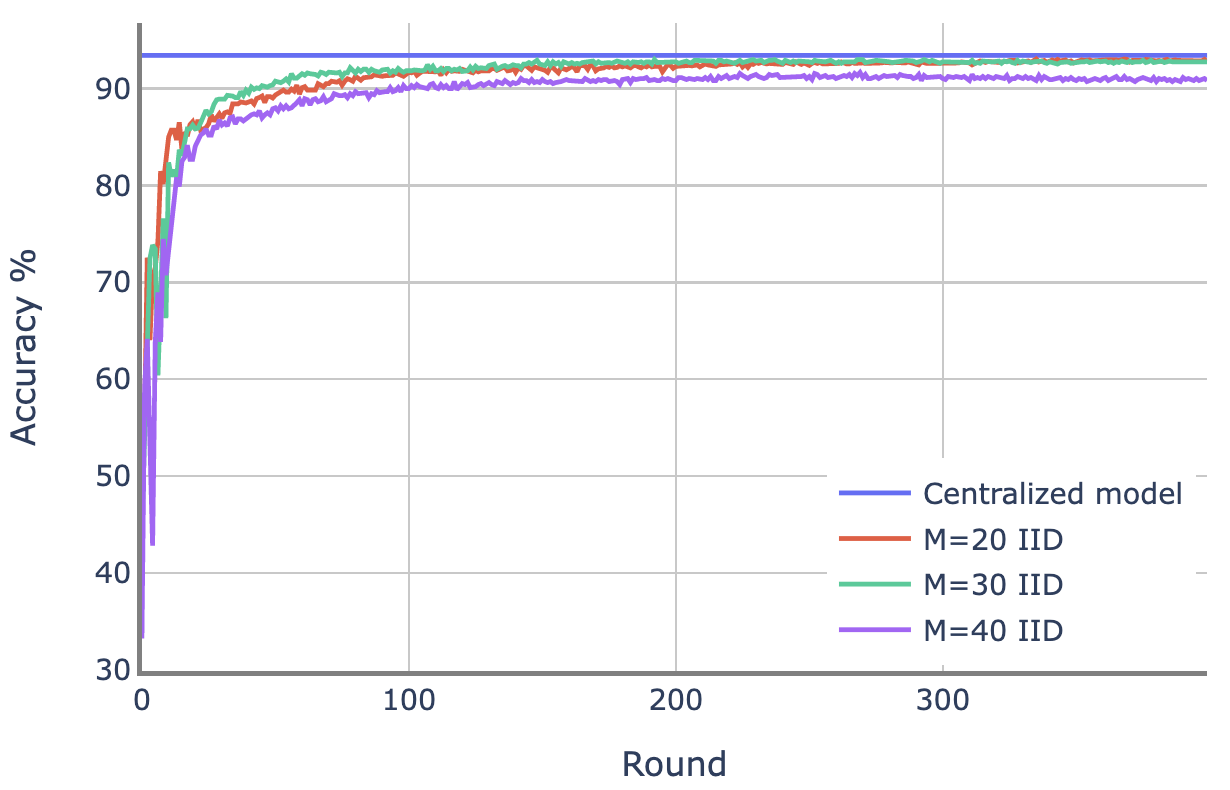}
}
\caption{Accuracy FedAvg in $400$ rounds under IID data setting. }
\label{fig:iid_accuracy}
\end{figure}



\begin{figure}
\centerline{\includegraphics[width=0.50\textwidth]{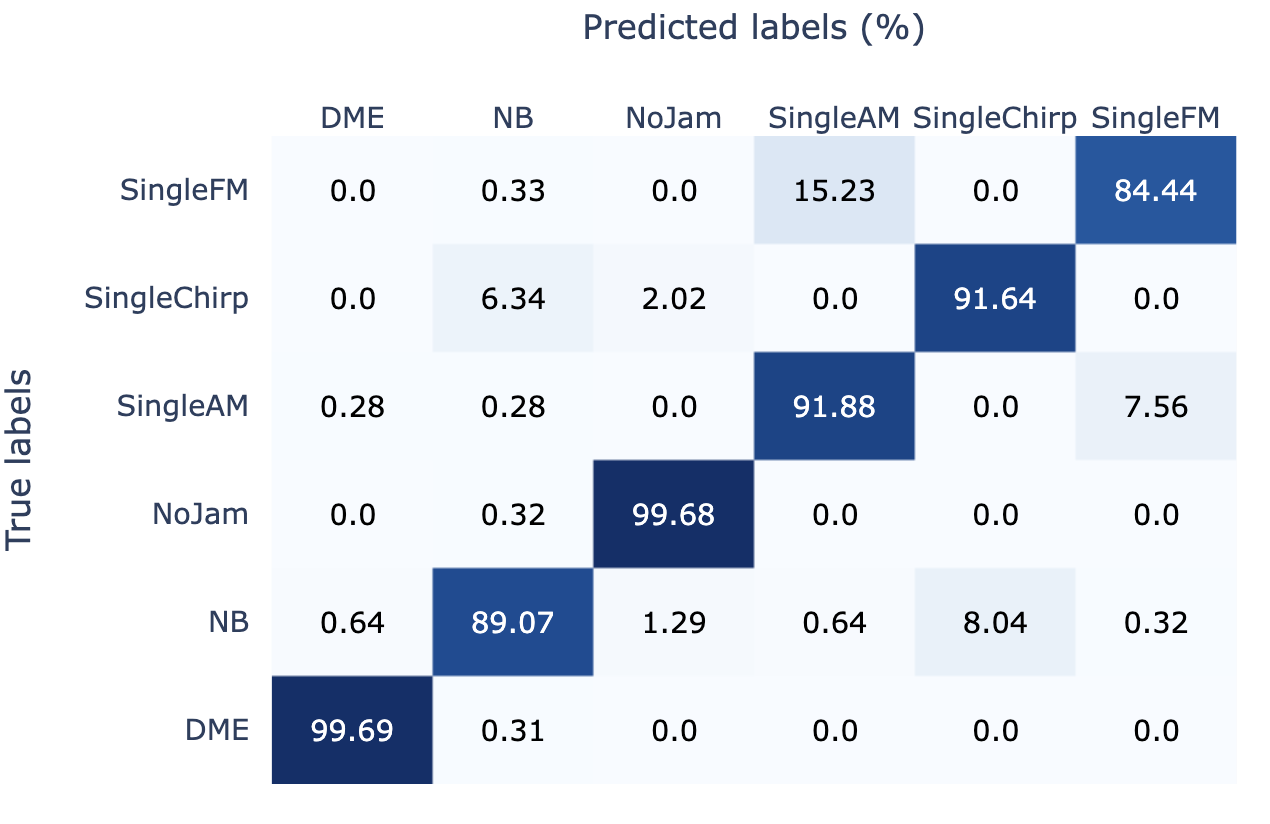}
}
\caption{Confusion matrix after FedAvg training under the IID data setting for $M=20$ clients.}
\label{fig:20_iid_cm}
\end{figure}

Figure \ref{fig:iid_accuracy} shows the accuracy of federated averaging algorithms with $400$ communication rounds under an IID data setting. The accuracy of the centrally trained model was used as benchmark (around 93.4\% accuracy). The figure also compares the accuracy when different numbers of clients $M$ were used. As expected, when a small number of clients were used, better results were achieved when compared to a larger number of clients. Intuitively, fewer clients have more data available and better train local models. Nevertheless, the results show high accuracy results for the tested number of clients. 

The confusion matrix in Figure~\ref{fig:20_iid_cm} reveals that each jammer class achieves relatively high accuracy, with the DME jammer type and the clean signal (i.e., type NoJam) providing the highest accuracies (over 99\%). The classifier is able to detect the absence of interference, as spectrogram (a) from Figure~\ref{fig:jammer_detect} notably differs from the rest. This is because the spectrum of a clean signal contains the signal of interest buried in Gaussian noise, which pollutes the whole spectrogram. On the other hand, as jamming signals are received with dramatically higher power than the satellite signal of interest, the noise $w(t)$ cannot be observed in spectrograms (b)-(f) from Figure~\ref{fig:jammer_detect}. Regarding DME (or pulsed) interferences, they are only active during their duty cycle. If the duty cycle is short with respect to the window duration of the short-time Fourier transform, the resulting spectrum shows an almost clean image with a few magnitude peaks, which notably differs from the spectra of other jammer types. The SingleFM and NB jammer types achieved less than 90\% accuracy. If inspecting the confusion matrix non-diagonal elements, it can be seen that it is difficult for the classifier to distinguish between the SingleAM and SingleFM types, as they all span one or two narrow bands of the signal spectrum. The SingleFM spectrogram is equivalent to the SingleAM spectrogram with an additional band. It is also difficult for the classifier to distinguish between the NB and SingleChirp interferences. A possible explanation is that both of them have a lower magnitude in their spectra due to being more spread. This makes their spectrogram images look blurry when compared to the ones from the SingleAM and SingleFM types.


\begin{figure}
\centerline{\includegraphics[width=0.50\textwidth]{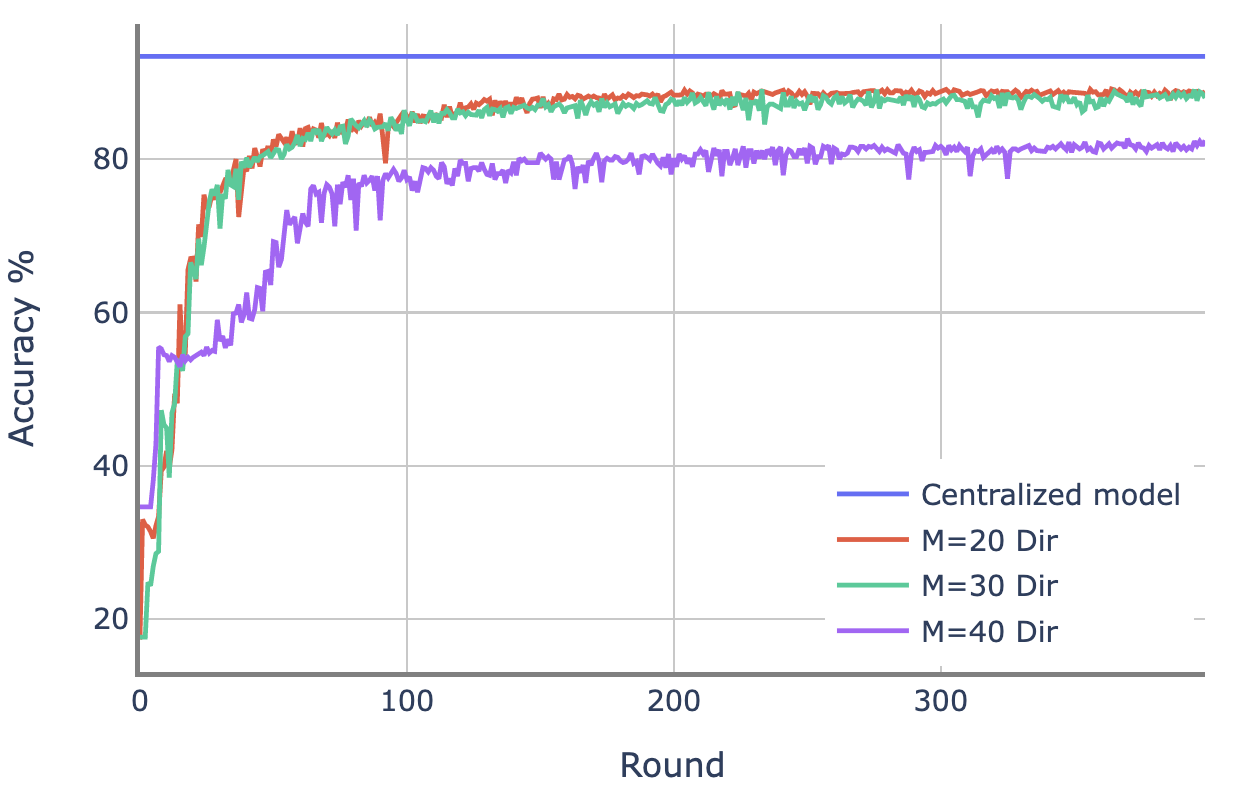}
}
\caption{Accuracy FedAvg in 400 rounds under Dirichlet data setting}
\label{fig:dir_accuracy}
\end{figure}



\begin{figure}
\centerline{\includegraphics[width=0.50\textwidth]{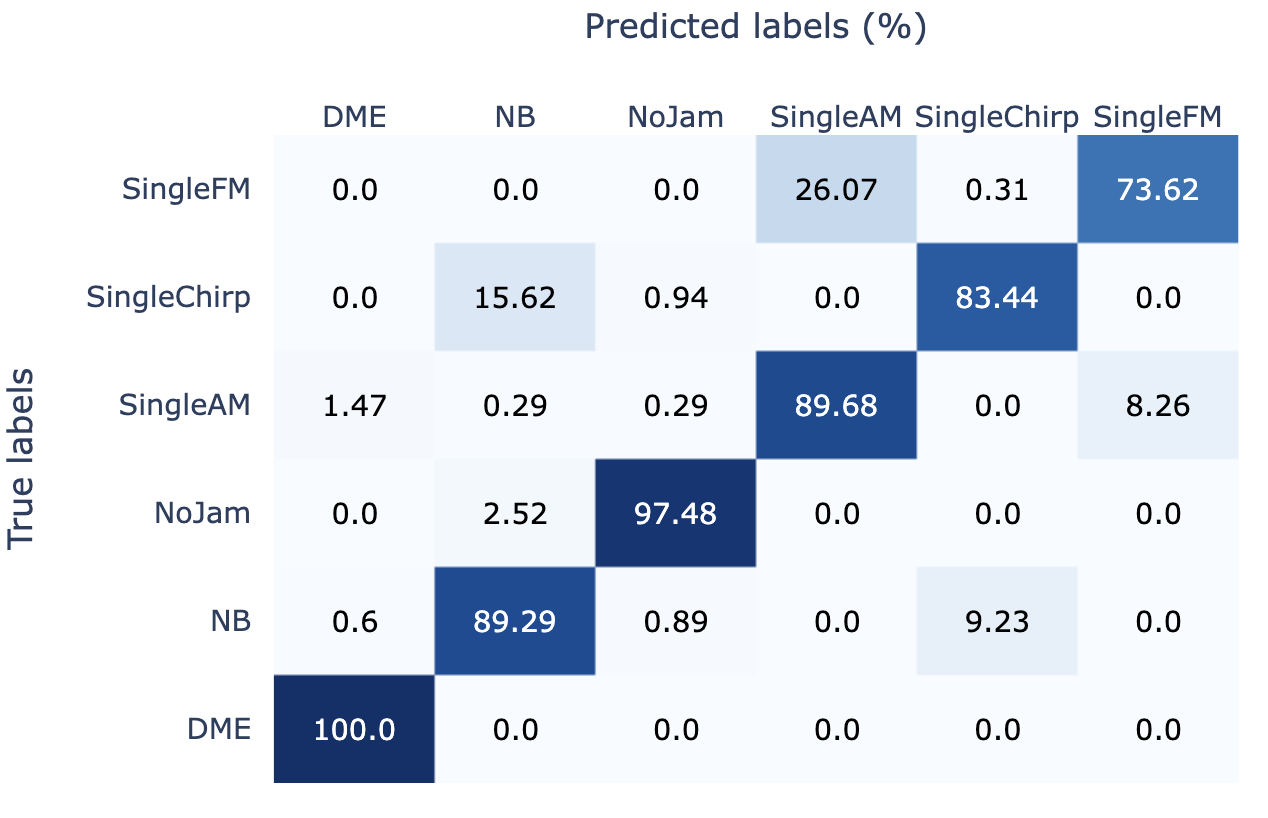}
}
\caption{Confusion matrix after FedAvg training under the non-IID data setting for $M=20$ clients.}
\label{fig:20_dir_cm}
\end{figure}

Figure \ref{fig:dir_accuracy} illustrates the accuracy of the FedAvg algorithm under a non-IID data setting, for different numbers of clients and compared to the accuracy of the global benchmark. The results show that the accuracies of the different client numbers are lower than the results of the homogeneous IID data setting, indicative of increased difficulty in learning with the heterogeneity of the data. Moreover, the comparison of different clients is similar to that of the IID data setting: the more clients there were, the lower their accuracy. It is also noticed that when the number of clients was $40$, it took more communication rounds to converge than when smaller numbers of clients were considered.


\begin{figure}
\centerline{\includegraphics[width=0.50\textwidth]{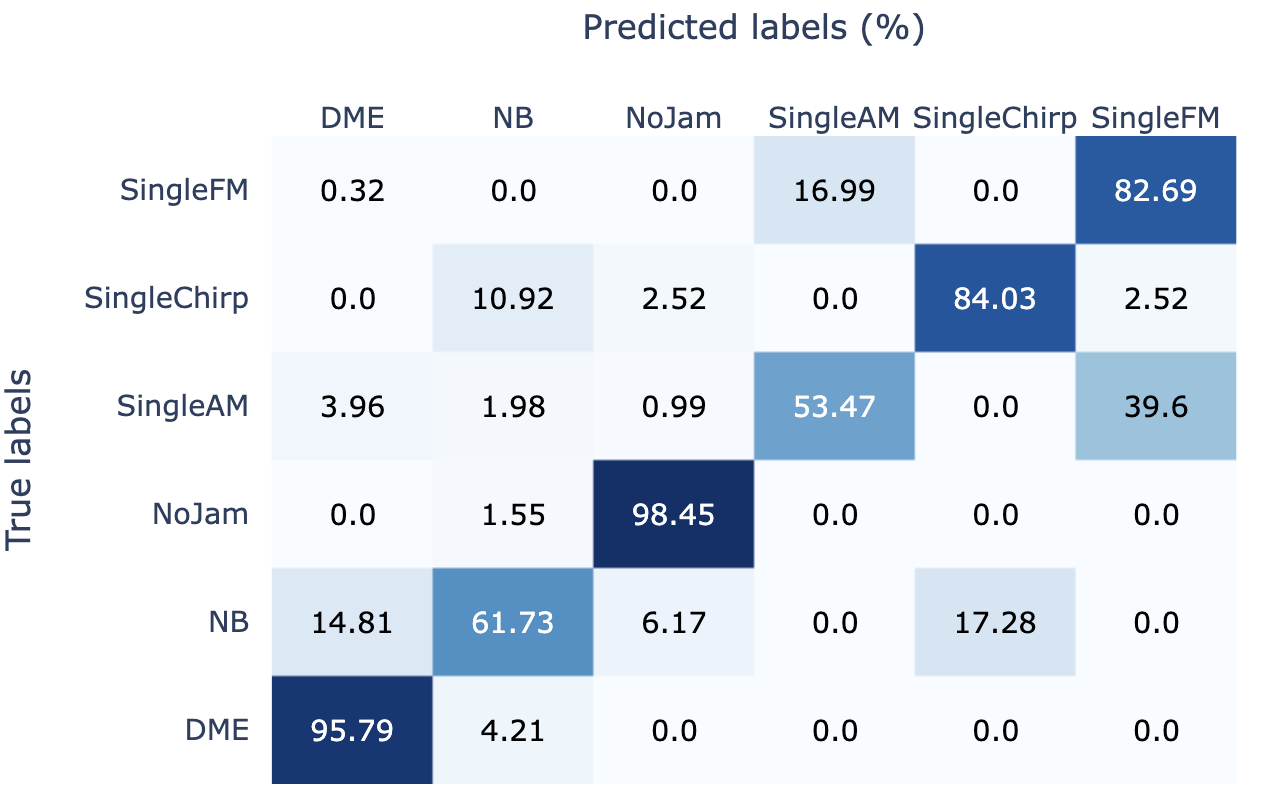}
}
\caption{Confusion matrix after FedAvg training under the non-IID data setting for $M=40$ clients.}
\label{fig:40_dir_cm}
\end{figure}

Figures~\ref{fig:20_dir_cm} and~\ref{fig:40_dir_cm} show the confusion matrix for $20$ and $40$ clients under the non-IID, Dirichlet data setting. The conclusion still holds that the DME jammer type and clean signal were the easiest to classify, with accuracies of 100\% and 97.48\% for $M=20$ and 95.79\% and 98.45\% for $M = 40$ clients. For $M = 40$, very low accuracies were achieved with the NB and SingleAM jammer types, while the worst accuracy was achieved with the SingleFM jammer type for $M = 20$. As in Figure~\ref{fig:20_iid_cm}, when inspecting the non-diagonal elements, it can be seen how it was difficult for the classifier to distinguish between SingleAM and SingleFM types, and also between NB and SingleChirp types. For $M = 40$, given that the performance was worse due to having a higher number of clients (implying less local data), it was also difficult for the classifier to distinguish between NB and DME signals. Nevertheless, for a high number of clients (i.e., $M=40$), accuracies above 80\% were obtained with the DME, clean signal, SingleChirp and SingleFM jammer types. For a lower number of clients (i.e., $M=20$), all jammer types could be classified with an accuracy above $80\%$.

As a final remark, the results presented in this section are comparable to the ones obtained with the benchmark training process: the centralized classification algorithm proposed in~\cite{Ferre2019}. In their results, the DME (or pulsed) interference and clean signal also provided the highest accuracy. Also, their confusion matrices showed the classifier difficulty when it comes to distinguishing SingleAM and SingleFM interferences, and also NB and SingleChirp interferences. Our obtained accuracies for $M = 20$ when classifying the DME and NB types exceed the accuracy provided by the benchmark CNN. Consequently, we have shown that the proposed federated learning framework allows to obtain comparable results to the ones offered by state-of-the art centralized classification algorithms while preserving user data privacy and security.

\section{Conclusion}\label{sec:conclusion}

This paper demonstrates the efficacy of FL in the context of GNSS jamming classification using the FedAvg, which would allow the successful implementation of a crowdsourcing scheme where real data is gathered without compromising user privacy. 
Results are provided for spectrogram image classification of simulated GNSS signal under the threat of six different jammer types.
%
%
Although classification accuracy results are high under certain configurations for all the studied jammer types, DME and clean signal provide the highest accuracies (above 99\%). On the other hand, it is difficult for the classifier to distinguish between AM and FM, as well as between NB and Chirp jammer types.
The FL framework performance has been successfully compared to the one provided by the benchmark centralized classification algorithm in~\cite{Ferre2019}, showing that it is possible to work in a collaborative scenario without observing a relevant performance drop while preserving user protection.
Experimental results showed that $i)$ it is more difficult to learn non-IID data than IID data; and that $ii)$
having fewer data on the local clients decreases the performance of the results. 

\balance
\bibliographystyle{IEEEtran}
\bibliography{biblio,t_trash,fed_antijamming}

\end{document}